%% file: acl2019.tex
\documentclass[11pt,a4paper]{article}
\usepackage[hyperref]{acl2019}
\usepackage{times}
\usepackage{latexsym}

\usepackage[utf8]{inputenc}
\usepackage{amssymb,amsfonts,amsmath}
\usepackage{graphicx}
\usepackage{multirow}
\usepackage{lipsum}
\usepackage{xcolor}

\usepackage{float}
\usepackage{wrapfig}
\restylefloat{figure}
\usepackage{booktabs}
\usepackage[english]{babel}

\usepackage{caption}
\usepackage[flushleft]{threeparttable}
\usepackage{subcaption}
\usepackage{footnote}
\makesavenoteenv{tabular}
\makesavenoteenv{table}
\usepackage{setspace}
\usepackage{natbib}
\usepackage{array}
\usepackage{calc}
\usepackage{url}

\newcommand{\x}{\mathbf{x}}
\newcommand{\y}{\mathbf{y}}

\newcommand{\p}{\mathrm{p}}

\newcommand{\VTheta}{\mathbf{\Theta}}

\DeclareMathOperator*{\argmax}{arg\,max~}

\definecolor{darkgreen}{RGB}{0,100,0}
\newcommand*{\regularbox}[1]{\color{black}% open a group for a local setting
	\setlength{\fboxsep}{-1\fboxrule}% the rule will be inside the box boundary
	\fbox{\hspace{1.1pt}\textbf{\strut#1}\hspace{1.2pt}}% print the box, with some padding at the left and right
	\color{black}  }% close the group
\aclfinalcopy % Uncomment this line for the final submission
\usepackage[capitalize]{cleveref}

\title{A Neural, Interactive-predictive System \\
	  for Multimodal Sequence to Sequence Tasks}

\author{\'{A}lvaro Peris \and Francisco Casacuberta \\
	Pattern Recognition and Human Language Technology Research Center \\ Universitat Politècnica de València, València, Spain\\
	\texttt{\{lvapeab, fcn\}@prhlt.upv.es}}

\date{}

\begin{document}
	\maketitle
	\begin{abstract}
	We present a demonstration of a neural interactive-predictive system for tackling multimodal sequence to sequence tasks. The system generates text predictions to different sequence to sequence tasks: machine translation, image and video captioning. These predictions are revised by a human agent, who introduces corrections in the form of characters. The system reacts to each correction, providing alternative hypotheses, compelling with the feedback provided by the user. The final objective is to reduce the human effort required during this correction process.

	This system is implemented following a client--server architecture. For accessing the system, we developed a website, which communicates with the neural model, hosted in a local server. From this website, the different tasks can be tackled following the interactive-predictive framework. We open-source all the code developed for building this system. The demonstration in hosted in \url{http://casmacat.prhlt.upv.es/interactive-seq2seq}.

	\end{abstract}
	
	\input{1_intro.tex}
	\input{3_system.tex}

\input{4_visualization.tex}
	\input{5_conclusions.tex}
	
	\section*{Acknowledgments} We acknowledge the anonymous reviewers for their helpful suggestions. The research leading to these results has received funding from the Generalitat Valenciana under grant PROMETEOII/2014/030 and from TIN2015-70924-C2-1-R. We also acknowledge NVIDIA Corporation for the donation of GPUs used in this work.
	
	\bibliography{acl2019}
	\bibliographystyle{acl_natbib}
	
	\appendix

\end{document}

%% file: 1_intro.tex
\section{Introduction}

The sequence to sequence problem involves the transduction of an input sequence $\x$ into an output sequence $\hat{\y}$ \citep{Graves12}. In the last years, many tasks have been tackled under this perspective using neural networks with extraordinary results: neural machine translation \citep[NMT;][]{Sutskever14,Bahdanau15}, speech recognition and translation \citep{Chan16,Niehues18}, image and video captioning \citep{Xu15,Yao15}, among others.

These systems are usually based on the statistical formalization of pattern recognition \citep[e.g.][]{Bishop06}. Following this probabilistic framework, the objective is to find most likely output sequence $\hat{\y}$, given an input sequence $\x$, according to a model $\VTheta$:

\begin{equation}
\label{eq:smt-eq}
\hat{\y} = \argmax_{\y} \p(\y \mid \x ; \VTheta)
\end{equation}

In the last years, $\VTheta$ has been frequently implemented as a deep neural network, trained in an end-to-end way. These neural systems have consistently outperformed other alternatives in the aforementioned problems. However and despite these impressive advances, the systems are not perfect, and still make errors \citep{Koehn17}. 

In several scenarios, and especially in machine translation, fully-automatic systems are usually used for providing initial predictions to the input objects. These predictions are later revised by a human expert, who corrects the errors made by the system. This is known as post-editing and, in some scenarios, it increases the productivity with respect to performing the task from scratch \citep{Alabau16,Arenas08,Hu16}. 

\begin{figure*}[!ht]
	\centering
	\includegraphics[width=0.9\textwidth]{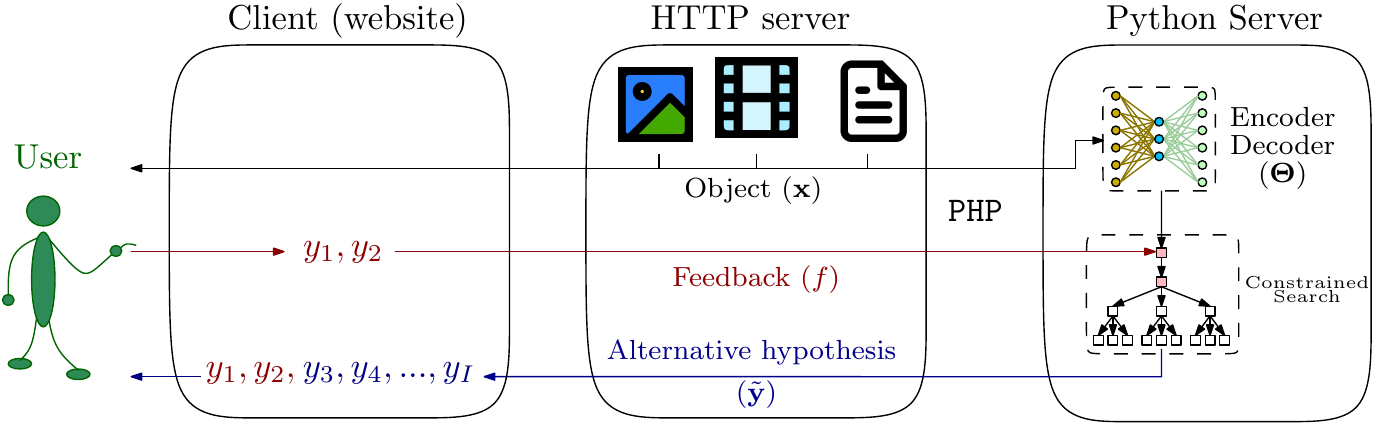}
	\caption{\label{fig:system-architecture}System architecture. The client, a website, presents the user several input objects (images, videos or texts) and a prediction. The user then introduces a feedback signal, for correcting this prediction. After being introduced, the feedback signal is sent to the server---together with the input object---for generating an alternative hypothesis, which takes into account the user corrections.}
\end{figure*}

\subsection{Interactive-predictive pattern recognition}

As an alternative to the static, decoupled post-editing, other strategies have been proposed, aiming to improve the productivity of the correction phase. Among them, the interactive-predictive pattern recognition \citep{Foster97} results particularly interesting. Under this framework, the static correction stage shifts to an iterative human-computer collaboration process.

The user interacts with the system by means of a feedback signal $f$. The system suggests then an alternative hypothesis $\tilde{\y}$, compatible with the feedback. The inclusion of the feedback into the general pattern recognition rewrites \cref{eq:smt-eq} introducing a restriction on the search space:

\begin{equation}
\label{eq:imt}
\tilde{\y}  = \argmax_{\y~\textrm{compatible with}~f} \p(\y \mid \x, f; \VTheta)
\end{equation}

The most paradigmatic application of the interactive-predictive pattern recognition framework is machine translation. The addition of interactive protocols to foster productivity of translation environments have been studied for long time, for phrase-based models \citep{Alabau13,Alabau16,Barrachina09,Federico14,Green14} and also for NMT systems \citep{Knowles16,Peris17a,Peris19,Wuebker16}. 

The system we are presenting in this work is an extended version of \citet{Peris19}, who presented a NMT system that accepted a prefix feedback: the user corrected the first wrong character of the sentence. Hence, the system reacted to the feedback by providing an alternative suffix. This protocol can be implemented as a constrained beam search. Moreover, the system can be retrained incrementally, as soon as a corrected sample is validated, following an online learning scenario.

We generalize this interactive-predictive NMT system to cope with alternative input modalities, namely images and videos. The system can be accessed following a client--server interface. We developed a client website, that access to our servers, in which the interactive-predictive systems are deployed. A live demo of the system can be accessed in: \url{http://casmacat.prhlt.upv.es/interactive-seq2seq}.

In the following sections, we describe the main architecture, features and usage of our interactive-predictive system. We also describe the frontend of our demonstration website and present an example of interactive session.

%% file: 3_system.tex
\begin{figure*}[!h]
	\centering
	\includegraphics[width=.95\textwidth]{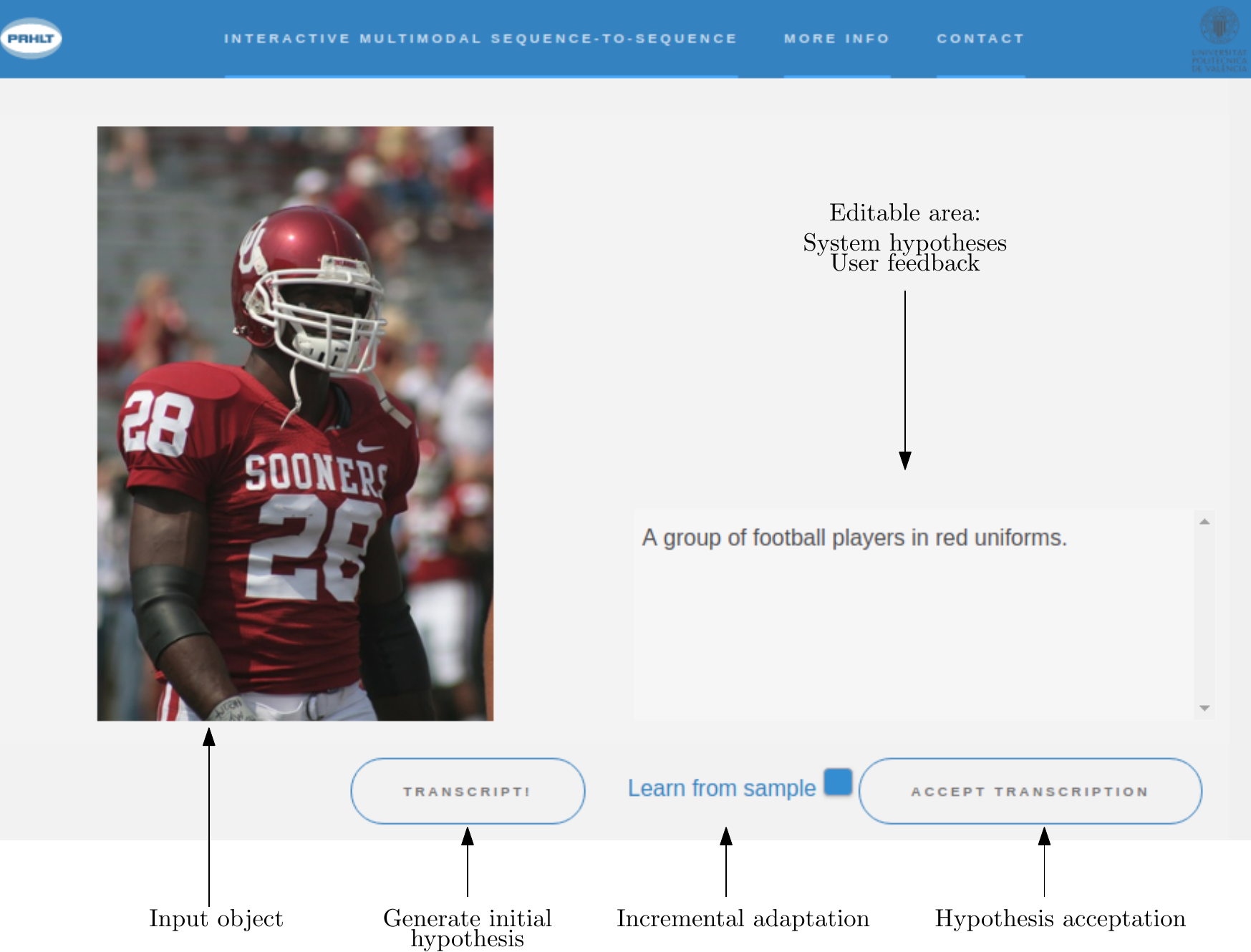}
	\caption{\label{fig:example}Frontend of the client website. As the button ``Transcript!'' is clicked, an initial hypothesis for the input object---in this case, an image---appears in the right area. The user then introduces corrections of this text. The system reacts to each translation, producing alternative hypotheses, always compliant with the user feedback. Once a correct caption of the image is reached, the user clicks in the ``Accept translation'' button, validating the hypothesis. }
\end{figure*}

\section{System description}

The core of our system is a neural sequence to sequence model, developed with NMT-Keras \citep{Peris18b}. This library is built upon Keras \citep{Chollet15} and works for the Theano \citep{Theano16} and Tensorflow \citep{Abadi16} backends. The system is deployed as a Python-based HTTP server that waits for requests. The user interactions are introduced through a (client) HTML website. The website is hosted on a Nginx server that manages the interactions using Javascript and communicates with the Python server, using the PHP curl tool. All code is open-source and publicly available\footnote{Python server source code: \url{https://github.com/lvapeab/interactive-keras-captioning}.}\footnote{HTML server source code: \url{https://github.com/lvapeab/inmt_demo_web}.}. 

NMT-Keras extends the (already extensive) Keras functionalities, providing a flexible, easy to use framework upon which build neural models. Among the features brought by NMT-Keras, some of them are particularly useful for sequence-to-sequence tasks: extended recurrent neural networks, with embedded attention mechanisms and conditional LSTM/GRU units \citep{Sennrich17}, multi-head attention layers, positional encodings and position-wise feed-forward networks for building Transformer models \citep{Vaswani17} and a modular handler for processing different data modalities, including text, images, videos or categorical labels. 

Within this framework, we built our neural systems, which are leveraged via our interactive client--server application. The neural systems are deployed in a server, waiting for requests. When the client ask for a prediction, they react, generate the prediction and deliver it back to the client.

\subsection{Usage of the interactive system}

\begin{figure*}[!ht]
	\centering
%	\def\arraystretch{1.1}
	%	\footnotesize
	\begin{tabular}{ccl}
		\toprule
		\textbf{0} & \textbf{System} & A group of football players in red uniforms.
		\\
		\midrule
		\multirow{2}{*}{\textbf{1}} & \textbf{User} & \textcolor{darkgreen}{\textit{A}} \regularbox{f}group of football players in red uniforms. \\
		& \textbf{System} &\textcolor{darkgreen}{ \textit{A f}}ootball player in a red uniform is holding a football. \\
		\midrule
		\multirow{2}{*}{\textbf{2}} & \textbf{User} & \textcolor{darkgreen}{\textit{A football player in a red uniform is }} \regularbox{w}holding a football. \\
		& \textbf{System} &\textcolor{darkgreen}{ \textit{A football player in a red uniform is w}}earing a football. \\
		\midrule
		\multirow{2}{*}{\textbf{3}} & \textbf{User} & \textcolor{darkgreen}{\textit{A football player in a red uniform is wearing a }} \regularbox{h}football. \\
		& \textbf{System} &\textcolor{darkgreen}{ \textit{A football player in a red uniform is w}}earing a helmet. \\
		\midrule
		\multirow{1}{*}{\textbf{4}} & \textbf{User} & \textcolor{darkgreen}{\textit{A football player in a red uniform is wearing a helmet.}} \\
		
		\bottomrule
	\end{tabular}
	
	\caption{\label{fig:exampleIMT} Interactive-predictive session for correcting the caption generated in \cref{fig:example}. At each iteration, the user introduces a character correction (boxed). The system modifies its hypothesis, taking into account this feedback: keeping the correct prefix (green) and generating a compatible suffix.
	}
\end{figure*}

Our interactive-predictive system works as follows: initially, an input object is presented to the user in the client website. The user requests an automatic prediction of it. Next, the client communicates the server via PHP. The server queries the neural system, which produces an initial hypothesis applying \cref{eq:smt-eq}. The hypothesis is then sent back to the client website.

Next, the interactive-predictive process starts: the user searches in this hypothesis the first error, and introduces a correction with the keyboard (writing one or more characters). When the user stops typing the correction, the system reacts, sending to the server a request containing the input object and the user feedback (the sequence of characters that conform the correct prefix). Then, the neural model implements \cref{eq:imt} and produces an alternative hypothesis, such that it completes the correct prefix. This is implemented as a constrained beam search, as described in \citet{Peris17a,Peris19}. This iteration of the process is illustrated in \cref{fig:system-architecture}.

This protocol is repeated until the user finds satisfactory the hypothesis given by the system. Then, it is validated. As soon as the sentence is validated, the system can be incrementally updated with this sample, following an online learning setup \citep{Peris19}. Hence, in future interactions, the system will be progressively updated, tailoring to a given domain or to the user preferences. These adaptive systems have shown to be effective for reducing the human effort spent in the process \citep{Karimova18}.

%% file: 4_visualization.tex
\section{System showcase}

To show the interactive-predictive protocol described in the previous sections, we developed a website which hosts a demonstration of the system. Our demonstration system handles three different problems, regarding three different data modalities: text-to-text (NMT), image-to-text (image captioning) and video-to-text (video captioning). For tackling these tasks, we use a similar model: a neural encoder--decoder, based on recurrent neural networks with attention \citep{Bahdanau15,Xu15,Yao15}. Our framework has also support for Transformer-like architectures \citep{Vaswani17}.

The NMT task regards the translation of texts from a medical domain. The system is similar to the one used by \citet{Peris19}, and was trained on the UFAL corpus \citep{Bojar17c}. The image and video captioning systems were trained on the Flickr8k \citep{Hodosh10} and MSVD \citep{Chen11} datasets, respectively. The images were encoded using an Inception convolutional neural network \citep{Szegedy16} trained on the ILSVRC dataset \citep{Russakovsky15}. The decoder receives the representation previous to the fully-connected work. In the case of the video captioning system, we applied a 3D convolutional neural network \citep{Tran15}, for obtaining time-aware features. 

Finally, as aforementioned in previous sections, the systems can be retrained after the validation of each sample. In our demonstration, the systems are updated via gradient descent, but using a learning rate of $0$, which prevents a degradation of the model due to accidental misuse.

\subsection{Example: image captioning}

We show and analyze an image captioning example. The NMT and video captioning tasks are similar. \cref{fig:example} shows the demo website, for the image captioning task. In the left part of the screen, the input object is shown, in this case, an image. As the user clicks in the ``Transcript!'' button, the system generates a caption of the image, displaying it in an editable area on the right part of the screen. The user can then introduce the desired corrections to this hypothesis. As a correction is introduced, the system reacts, providing an alternative caption, but always considering the feedback given by the user.

As can be seen in \cref{fig:example}, the caption generated by the system has some errors. \cref{fig:exampleIMT} an shows the interactive-predictive captioning session, for obtaining a correct sample. With three interactions, the system was able to obtain a correct caption for the image. 

It is particularly interesting to observe that the system correctly accounts for the singular/plural concordance of the clause \textit{in red uniform(s)}, depending on the subject (\textit{A football player}/\textit{A group of football players}).

%Malos ejemplos

%% file: 5_conclusions.tex
\section{Conclusions and future work}

We presented a demonstration of a interactive-predictive neural system for multimodal sequence to sequence tasks. We described its client--server architecture and developed a website for ease the usage of the system.

As future work, we would like to improve the frontend of our website. Inspecting the attributes of black-box neural models is a relevant research topic, and it is under active development \citep[e.g.][]{Zeiler14,Ancona17}. Visualizing these relevant attributes would help to understand the model predictions and behavior. 

Moreover, a more sophisticated frontend would allow to implement interesting features, such as mapping the attention weights through the input sequence or the implementation of more complex interaction protocols, such as touch-based interaction \citep{Marie15} or segment-based interaction \citep{Peris17a}. We intend to offer the different functionalities of the toolkit as REST services, for improving the reusability of the code. It is also planned to release the library in a Docker container in order to ease the deployment of future applications.